\documentclass[twocolumn, switch]{article} 

\usepackage{preprint}
\usepackage{algorithm}
\usepackage{algpseudocode}
\usepackage{amsmath, amsthm, amssymb, amsfonts}

\usepackage[numbers,square]{natbib}
\usepackage[utf8]{inputenc}	
\usepackage[T1]{fontenc}	
\usepackage[table]{xcolor}
\usepackage{colortbl}
\usepackage{pifont}
\definecolor{color_blue}{RGB}{135,206,235}
\definecolor{color_pink}{RGB}{252,182,165}
\definecolor{color_org}{RGB}{255,217,178}
\definecolor{color_yellow}{RGB}{255,255,204}
\definecolor{Gray}{gray}{0.95}
\definecolor{mygray}{gray}{.9}
\definecolor{r3}{RGB}{0,0,0}
\definecolor{cc}{RGB}{0,0,0}
\newcommand{\cmark}{\ding{51}}
\newcommand{\xmark}{\ding{55}}
\newcommand{\red}[1]{{\color{red}#1}}

\usepackage{graphicx}
\usepackage{eso-pic}
\usepackage[colorlinks = true,
            linkcolor = purple,
            urlcolor  = blue,
            citecolor = cyan,
            anchorcolor = black]{hyperref}	
\usepackage{booktabs} 		
\usepackage{nicefrac}		
\usepackage{microtype}		
\usepackage{lineno}		
\usepackage{float}			
\usepackage{placeins}

\usepackage{lipsum}		

\usepackage{newfloat}
\DeclareFloatingEnvironment[name={Supplementary Figure}]{suppfigure}
\usepackage{sidecap}
\sidecaptionvpos{figure}{c}

\usepackage{titlesec}
\titlespacing\section{0pt}{12pt plus 3pt minus 3pt}{1pt plus 1pt minus 1pt}
\titlespacing\subsection{0pt}{10pt plus 3pt minus 3pt}{1pt plus 1pt minus 1pt}
\titlespacing\subsubsection{0pt}{8pt plus 3pt minus 3pt}{1pt plus 1pt minus 1pt}

\title{Towards Training-free Automatic Proxy Discovery via Large Language
Models for Mixed Precision Quantization}

\usepackage{eso-pic}
\usepackage{tikz}
\usepackage[table]{xcolor}
\PassOptionsToPackage{colorlinks=true,linkcolor=gray,urlcolor=gray}{hyperref}

\usepackage{titling}
\usepackage{orcidlink}
\usepackage{footmisc}
\newcommand{\Author}[3]{
  \textbf{#1}\textsuperscript{#2},\ \orcidlink{#3} %
}

\author{
  \Author{Haidong Kang}{1*}{0000-0001-8533-5704} \and
  \Author{Jun Du}{2*}{0009-0007-0383-0631}\and
  \Author{Guo Yu}{3}{https://orcid.org/0000-0003-2427-8202}
}

\date{%
  \textsuperscript{1}School of Computer and Communication Engineering,
Northeastern University\\
  \textsuperscript{2}School of Software Engineering, Beijing
Jiaotong University\\
  \textsuperscript{3}Institute of Intelligent Manufacturing, Nanjing Tech University\\
  }

\begin{document}

\twocolumn[ 
  \begin{@twocolumnfalse} 

\maketitle
\thispagestyle{plain}

  \end{@twocolumnfalse} 
] 

\begin{abstract}
Mixed-Precision Quantization (MPQ) liberates the Deep Neural Networks (DNNs) from the Out-Of-Memory (OOM) bottleneck, which garnered increasing research attention. However, conventional methods either searched from costly differentiable optimization, which is neither efficient nor flexible, or learned a quantized DNN from the proxy (i.e., HAWQ) manually designed by human experts, which is labor-intensive and requires huge expert knowledge. Can we design a proxy without involving any human experts and training? In this paper, we provide an affirmative answer by proposing a novel Large Language Models (LLMs)-driven Training-free Automatic Proxy (dubbed, TAP) discovery framework, which reforms the design paradigm of MPQ by utilizing LLMs and evolutionary search strategies to find superior TAP tailored for MPQ, automatically. In addition, to bridge the gap between black-box LLMs and the tough MPQ task, we introduce a lightweight Direct Preference Optimization (DPO)–based strategy controller that dynamically reweights the selection probabilities of the three prompt templates for evolutionary search strategies according to fitness signals, without fine-tuning the LLM. This builds a task-aware feedback loop that improves proxy generation across evolutions. Extensive experiments on mainstream benchmarks demonstrate that TAP achieves state-of-the-art performance. Finally, we truly believe that our TAP will significantly contribute to the MPQ community by providing a new perspective on LLM-driven design algorithms.
\end{abstract}



\section{Introduction}
\label{sec_introduction}



Deep neural networks have become indispensable in modern vision applications. However, when deployed on extremely resource-limited devices such as MCUs and tiny NPUs, they often encounter severe Out-Of-Memory (OOM) issues due to the gap between model complexity and the tiny on-chip memory. This tension motivates the development of efficient compression techniques. Although fixed-precision quantization (FPQ) \cite{choi2018pact} provides a straightforward solution, its uniform bit assignment usually leads to unsatisfactory accuracy under strict hardware budgets. Mixed-precision quantization (MPQ) \cite{wang2019haq,dong2019hawq,yao2021hawq}, in contrast, adjusts bit-widths according to layer sensitivity and thus achieves a better balance between accuracy and efficiency. Existing differentiable MPQ methods \cite{cai2020rethinking,zhang2021differentiable,chu2021mixed,yang2021fracbits} further attempt to learn bit allocations automatically, yet their large computational cost severely limits practical deployment (Table \ref{tb:ob03}). As a response, training-free MPQ methods (e.g., HAWQ \cite{dong2019hawq}, HAWQ-V2 \cite{6}, OMPQ \cite{ma2023ompq}) avoid huge training costs, although they still require considerable manual effort.

\begin{table}[!t]
\centering
\caption{Performance comparison on various datasets.}
\label{tb:ob03}
\resizebox{0.45\textwidth}{!}{
\begin{tabular}{ccccc} 
\cline{1-5}
\textbf{Methods}         & \textbf{Top-1 (\%)} $\uparrow$                & \textbf{Bit (W/A)}   & \textbf{BOPs (G)} $\downarrow$ & \textbf{Cost.} $\downarrow$                      \\ 
\cline{1-5}
\multicolumn{5}{c}{\cellcolor{color_blue}{\textbf{ResNet-18}}} \\
\cline{1-5}
EdMIPS-Img \cite{cai2020rethinking}                & 65.9                                          & $\ast$/$\ast$        & 34.7                           & 9.5                                        \\
EdMIPS-C \cite{cai2020rethinking}              & 59.1                                          & $\ast$/$\ast$        & 7.4                            & 0.6                                          \\
\cline{1-5}
\multicolumn{5}{c}{\cellcolor{color_org}{\textbf{MobileNetV2}}} \\
\cline{1-5}
HAQ-Img \cite{wang2019haq}                         & 71.5                                          & $\ast$/32            & 42.8                           & 51.1                                        \\
HAQ-C \cite{wang2019haq}                       & 62.7                                          & $\ast$/32            & 8.1                            & 4.5                                             \\
\bottomrule
\end{tabular}}
\end{table}

\noindent{\bf Challenges.}
Although these training-free approaches have moved the field forward, they still face two essential challenges.
First, \textit{they remain strongly dependent on handcrafted heuristics}. As summarized in Table \ref{tab:review_zc}, existing proxies are constructed through expert knowledge, such as Hessian-based analysis in HAWQ or manually selected weight–activation statistics in OMPQ. Designing such rules usually involves extensive trial-and-error, which not only increases labor cost but also makes it difficult to adapt the proxies to new architectures and hardware constraints.
Second, Strong dependence on calibration data and slow convergence.
These expert-crafted proxies typically require large calibration sets and many optimization iterations. For example, HAWQ-V2 needs 8,192 samples and 50 iterations, and HAWQ involves more than 2,500 updates; even lightweight OMPQ still relies on nontrivial calibration and tuning. In contrast, our TAP needs only 16 samples and 5 iterations, revealing the inefficiency and instability of existing methods. These observations indicate that manual proxy design has reached a practical bottleneck and calls for a more scalable alternative.

\begin{table*}[!t]
\caption{TAP \textit{v.s.} previous methods. $H$, $A$, and $W$ denote the Hessian matrix, activation, and weight parameters, respectively.}
\centering
\resizebox{0.95\linewidth}{!}{
\begin{tabular}{ccccccc}
\hline
\textbf{Name} & \textbf{Formula} & \textbf{Human Expert} & \textbf{Calibration Data} & \textbf{Calibration Sample$\downarrow$} & \textbf{Weight Update} & \textbf{Convergence Steps$\downarrow$} \\ \hline
HAWQ \cite{dong2019hawq} &  $\begin{aligned}\label{eqa:hawqv1}\max_i \{\lambda_i(H)\} \end{aligned}$   &  \cmark &  \cmark & - & \xmark & 2500\\
HAWQ-V2 \cite{6} & $\begin{aligned}\label{eqa:hawqv2}\frac{1}{n}\sum_{i=1}^n \mathit{tr}(H_i)\end{aligned}$  & \cmark & \cmark & 8192 & \xmark & 50 \\
OMPQ \cite{ma2023ompq} & $\begin{aligned}\label{eqa:ompq}\frac{||z_j^T z_i||^2_{z}}{||z_i^T z_i||^2_{z} ||z_j^T z_j||^2_{z}}\end{aligned}$  & \cmark & \cmark & 64 & \xmark  &  -\\  \hline
\cellcolor{mygray}{\textbf{TAP}} &
\cellcolor{mygray}{$\|\mathbf{W}_{ij}\|_2 \cdot H\left( \mathbf{A}_{ij} \right) \cdot e^{-\frac{d_i}{L}}$} &
\cellcolor{mygray}{\xmark} &
\cellcolor{mygray}{\cmark} &
\cellcolor{mygray}{\textbf{16}} &
\cellcolor{mygray}{\xmark} &
\cellcolor{mygray}{\textbf{5}} \\
\hline
\end{tabular}}
\label{tab:review_zc}
\end{table*}

\noindent{\bf Motivations.}
Unlike previous training-free MPQ methods, Table \ref{tab:review_zc} suggests that there exists a promising direction: automatically discovering MPQ proxies with the help of Large Language Models (LLMs). This shifts proxy construction from expert heuristics to automated reasoning, thereby reducing manual effort while alleviating the dependence on heavy calibration.

\noindent{\bf Contributions.} In this work, we aim to analyze and address the above limitations. We first revisit the design principles of existing proxies and empirically verify their drawbacks in terms of expert dependency, calibration cost and convergence behavior (Section \ref{sec:Preliminaries}). Motivated by the advancement of LLMs in generating structured knowledge and reasoning \cite{radford2018improving,guo2025deepseek}, we propose TAP, the first framework that automatically constructs training-free MPQ proxies. At the beginning, we examine a simple prompting strategy. However, similar to the observations in other automated tasks, directly prompting the LLM yields unstable layer-wise bit allocations. A closer analysis reveals that this is largely due to the absence of any feedback signal that connects the generated proxy to its MPQ performance. To overcome this issue, we draw inspiration from the reasoning behaviors observed in Qwen3 \cite{qwen3} and DeepSeek \cite{guo2025deepseek}, and introduce a lightweight Direct Policy Optimization (DPO) \cite{rafailov2024directpreferenceoptimizationlanguage} reinforcement learning procedure. This introduces task-aware feedback into the prompt evolution, encouraging the LLM to produce intermediate reasoning steps before generating the final proxy. Thereby, the proxy quality improves steadily through iterative refinement. Our contributions are summarized as follows:

\begin{itemize}
\item \textbf{A new proxy-design paradigm for MPQ.}
We present TAP, an LLM-driven framework that adopts evolutionary search strategies and automatically discovers training-free proxies for mixed-precision quantization without relying on handcrafted rules, thus offering a fresh perspective on the design of mixed-precision quantization.

\item \textbf{DPO as a template selector.}
We identify the core weakness of naive prompting strategies and introduce a direct preference optimization (DPO)-based strategy controller that reallocates selection probabilities for three prompt templates according to fitness signals—without updating large language model (LLM) parameters—resulting in more reliable reasoning during proxy generation and superior integrated accuracy–efficiency performance.

\item \textbf{Extensive empirical validation.}
Experiments on mainstream benchmarks demonstrate that TAP achieves superior performance over existing expert-designed proxies, while being significantly more efficient in terms of calibration and convergence.
\end{itemize}

\section{PRELIMINARIES and Motivations}
\label{sec:Preliminaries}
In what follows, we first formulate the problem of MPQ. Next, we summarize how DMPQ optimizes learnable bit-width parameters ($\alpha$). Finally, we offer key observations of DMPQ and analyze their root causes via an in-depth analysis that facilitates MPQ exploration for target hardware.

\subsection{Rethinking the Training-free MPQ}
\noindent{\textbf{Problem Definitions}}.  
Given a network $F$ with layers $\{L_l\}_{l=1}^n$, training-free MPQ assigns bit-widths without optimization. Instead of solving the ERM objective, such methods construct a proxy:
\begin{equation} \centering \begin{aligned} \begin{matrix}s_l=\Phi_l(W_l,A_l),&\\\end{matrix} \end{aligned} \label{eq1} \end{equation}

And determine $(b_l^w,b_l^a)$ directly from the sensitivity ranking implied by $\{s_l\}$. The resulting quantized model is:
\begin{equation} \centering \begin{aligned} \begin{matrix}F^{q}(x)=\sum_{l=1}^{n} Q(F_l; b_l^w, b_l^a).&\\\end{matrix} \end{aligned} \label{eq2} \end{equation}

Thus, the core of training-free MPQ reduces to designing an effective proxy $\Phi_l$.



\subsection{Observations and Motivations}
\label{sec:Preliminary}
As shown in Table~\ref{tab:review_zc}, we can clearly observe that existing proxies (i.e., HAWQ, OMPQ) for training-free MPQ still depend on hand-crafted rules (e.g., Hessian traces or activation–weight statistics), which results in huge labor costs. Moreover, those methods suffer from large calibration sets, such as HAWQ-V2 requiring $8{,}192$ samples and $50$ iterations, and HAWQ over $2{,}500$ updates. Even OMPQ relies on manual patterns, making these methods costly and sensitive to expert design. For a new task or dataset, MPQ proxies face a dilemma: they must either be re-searched from scratch, which is computationally expensive and inflexible, or transferred from proxies designed for other tasks, which risks sub-optimal quantization performance due to task mismatch. Motivated by the reasoning abilities of LLMs, we investigate whether they can automatically generate effective training-free proxies for MPQ. Using a simple prompt (\textbf{App. \red{A}}), LLMs can produce proxy candidates, but naive prompting yields inconsistent performance due to the lack of task-aware feedback. To overcome this, we adopt a Direct Preference Optimization (DPO)–based strategy controller that reweights three prompt templates using fitness feedback, without fine-tuning the LLM. This establishes a task-aware feedback loop that iteratively refines proxy quality with lightweight signals. As a result, LLM-driven proxy generation reduces dependence on experts and calibration, while producing more accurate and hardware-aware proxies, offering a scalable and efficient paradigm for training-free MPQ. Importantly, the TAP proxy metric presented in Table~\ref{tab:review_zc} is only one example among multiple optimal proxy-metric formulas generated by the large language model. This particular formula reorganizes and quantifies existing heuristic rules using mathematical constructs such as norms, entropy, and exponential decay. By drawing on vast domain knowledge, the LLM can extract and recombine effective heuristics; the resulting algorithms inherit the rationality of human expertise while overcoming its inherent limitations.

\section{Approach}
\subsection{Overview of TAP}
\begin{figure*}[htbp]
	\centering
	\includegraphics[width=\linewidth]{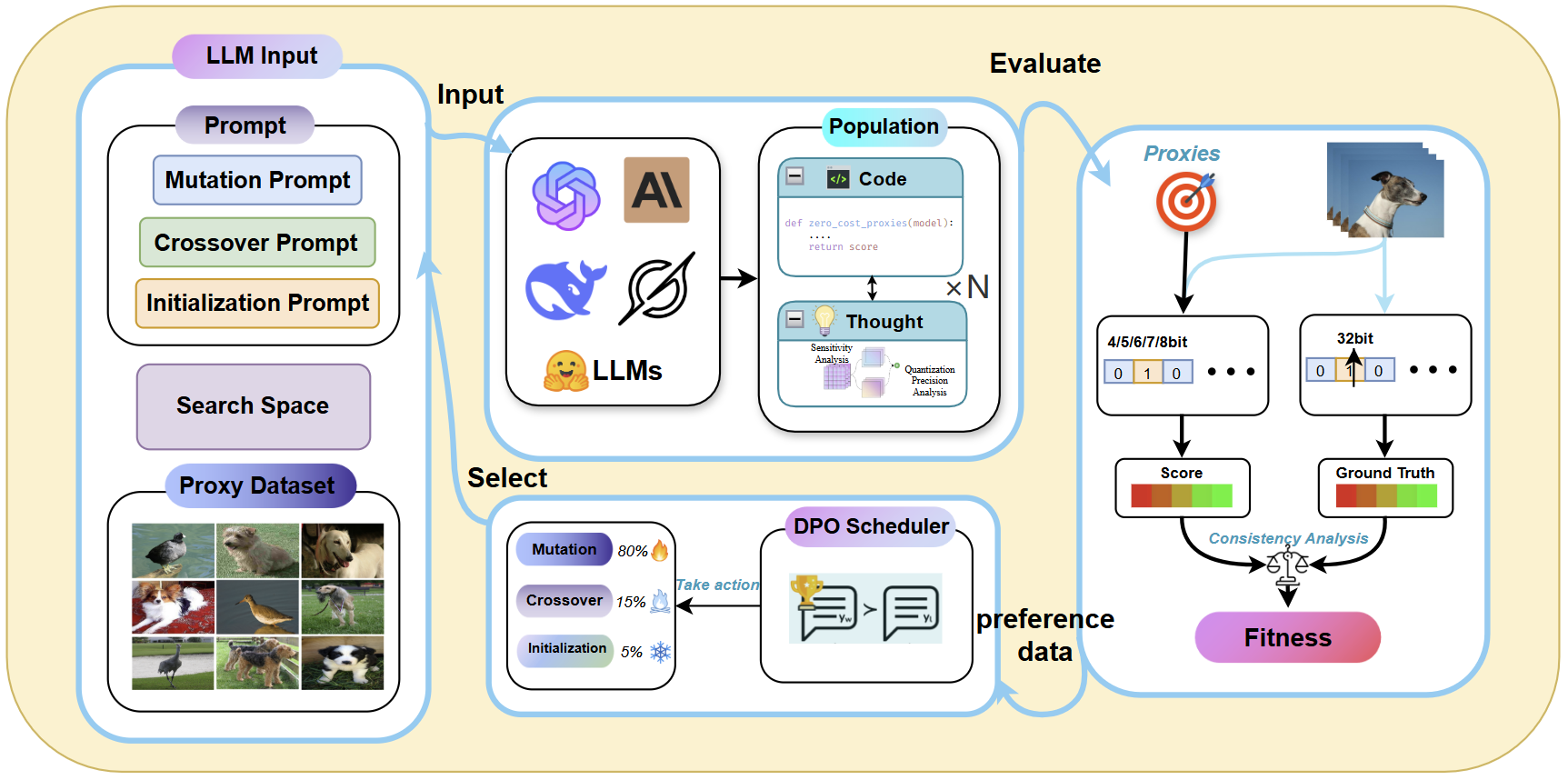}
	\caption{Overview of TAP. Input the initial prompt first. The LLM generates an initial proxy population, where each proxy must include inference logic, executable code, and bit-width allocation. Candidate proxies are validated on benchmarks like ImageNet-1k, with quantization accuracy and computational cost calculated to produce an adaptability score. Using this score, construct proxy preference data pairs and use a DPO-based strategy controller to dynamically adjust the selection probabilities of three prompt templates (without fine-tuning the LLM) for subsequent proxy generation. Merge the current and candidate populations, and filter by adaptability score to maintain a fixed size.}
	\label{overview}
\end{figure*}
To achieve the goal of designing zero-cost proxies for mixed-precision quantization tasks, the TAP framework leverages large language models and evolutionary search strategies to automatically generate proxies by optimizing natural language descriptions and their corresponding code. Furthermore, we propose an evolution scheduler based on Direct Preference Optimization (DPO), which acts as a non-parametric strategy controller that reallocates the selection probabilities of the three prompt templates for evolutionary search strategies according to fitness signals, without any fine-tuning of the large language models (LLMs). The overall architecture of the TAP framework is illustrated in Fig. \ref{overview}, consisting of three core components:

\noindent{\textbf{Proxy Candidate Generator.}} 
In the TAP framework, large language models (LLMs) serve as proxy candidate generators. Guided by meticulously designed prompts, LLMs can synthesize new zero-cost proxies or optimize existing ones, continuously generating higher-quality candidate proxies. Unlike traditional automatic proxy generation pipelines or complex task-specific algorithmic generators, our approach leverages LLMs to construct a vast implicit search space. Through adaptively optimized search strategies, this space can be systematically explored and incrementally refined, thereby approximating the optimal set of proxies.

\noindent{\textbf{Fitness Evaluator.}} 
The fitness evaluator rapidly quantifies the performance of each candidate proxy by computing the Spearman correlation between the candidate proxy and the actual accuracy on given mixed-precision quantization benchmarks (e.g., ImageNet-1k). Subsequently, the evaluator combines this statistical metric with population-level indicators such as population diversity and novelty to generate a comprehensive score. This score guides the evolutionary search process: on one hand, it maintains the information richness and balance of the proxy population; on the other hand, it eliminates low-utility candidate proxies.

\noindent{\textbf{RL Evolution Scheduler.}} 
To enable the proxy evolutionary strategy to be learnable and efficiently converge to optimal zero-cost proxies, the TAP framework incorporates a lightweight Direct Preference Optimization (DPO) module that serves as an evolutionary decision-maker. This DPO module does not require explicit reward function design; instead, it takes the "preference data pairs" (constructed by comparing the fitness scores of different proxies returned by the evaluator) as supervision signals and updates the selection weights of three prompt templates. Importantly, the LLM parameters remain fixed throughout; DPO only adjusts template selection probabilities to make the generated candidate proxies more aligned with the task's preference for "high performance and efficiency."

Specifically, given a search space \(\mathcal{A}\) of candidate bit-width configurations and their ground-truth performance \(p(a)\), the objective of the TAP framework is to learn a proxy \(f: \mathcal{A} \to \mathbb{R}\) whose scores preserve the performance ranking induced by p. Therefore, we aim to maximize the expected correlation:
\begin{equation}
\centering
    \begin{aligned}
        \max _{f \in \mathcal{F}} \mathbb{E}_{a \subseteq \mathcal{A}}[\rho(f(a), p(a))],
    \end{aligned}
    \label{eq01}
\end{equation}
where \(\rho\) denotes either Kendall’s \(\tau\) or Spearman’s \(\rho_s\) coefficient. Each proxy f is represented as a tuple \((T, C)\), where T is a natural language reasoning process describing the proxy’s underlying principle, and C is executable code that returns a scalar score. This representation ensures both interpretability and deterministic reproducibility of the proxy throughout the evolutionary process. For a subset of bit-width configurations \(a = (a_1, a_2, \cdots, a_n)^T\), we denote \(f(a)\) as \((f(a_1), f(a_2), \cdots, f(a_n))^T\); similarly, \(p(a) = (p(a_1), p(a_2), \cdots, p(a_n))^T\).

\subsection{Automatic Proxy Discovery for Mixed-Precision Quantization}
\label{subsec:automatic_proxy_discovery}

\noindent{\textbf{Proxy Candidate Generator.}} 
Let \(\mathcal{P}\) denote the set of valid proxies that satisfy the fixed input-output contract for mixed-precision quantization tasks. Specifically, each proxy must generate \textit{sensitivity scores} for convolutional layer channels and linear layers of neural networks—these scores directly guide the bit-width allocation process. At time step \(t\), given the current proxy population \(\mathcal{P}_t \ (\mathcal{P}_t \subseteq \mathcal{P})\), the Large Language Model (LLM) receives two types of inputs simultaneously: 
1. A structured prompt \(\pi \in \Pi\) specifying the operation type (initialization, mutation, crossover); 
2. A bounded context window containing existing proxies, including their natural language reasoning logic, executable code for sensitivity scoring, and historical bit-allocation performance. 
This input pair induces a context-conditioned distribution:
\begin{equation}
\mu_{\pi, \mathcal{C}_t} = P_{\mathcal{L}} \big( f \mid \pi, \mathcal{C}_t \big),
\label{eq:mu_context}
\end{equation}
where \(\mu_{\pi, \mathcal{C}_t}\) represents the probability distribution of candidate proxies generated by the LLM \(\mathcal{L}\) under the condition of prompt \(\pi\) and context \(\mathcal{C}_t\). By composing all prompts and admissible contexts, an implicit, context-aware search space is obtained:
\begin{equation}
\mathcal{F} = \bigcup_{t \in \mathbb{N}} \bigcup_{\pi \in \Pi} \bigcup_{\mathcal{C} \subseteq \mathcal{P}_t} \mathrm{supp}\big( \mu_{\pi, \mathcal{C}} \big),
\label{eq:F_space}
\end{equation}
where \(\text{supp}(\mu_{\pi, \mathcal{C}})\) denotes the support set of the distribution \(\mu_{\pi, \mathcal{C}}\), and \(\mathcal{F} \subseteq \mathcal{P}\). This structure allows the generator to reuse effective sensitivity scoring patterns and bit-allocation heuristics discovered in previous iterations.

\noindent{\textbf{Search Space and Constraints.}}
The proposed method performs an evolutionary search entirely within two coupled discrete spaces that are instantiated and navigated via prompt templates, without any first- or second-order model information:
- \textit{Proxy space} \(\mathcal{F}\). Each candidate proxy \(f \in \mathcal{F}\) is represented as a tuple \((T, C)\), where \(T\) is a natural-language reasoning trace and \(C\) is deterministic, executable code that produces a scalar score used to rank mixed-precision configurations. Proxies adhere to a fixed input–output contract and may only use architecture- and task-level metadata that do not require accessing internal training signals. In particular, the proxy design explicitly excludes the use of Hessian matrices, gradients, or raw weight/activation tensors.
- \textit{Configuration space} \(\mathcal{A}\). Each configuration \(a \in \mathcal{A}\) is a vector of per-layer bit-width assignments subject to hardware or compression constraints. In experiments, \(\mathcal{A}\) is instantiated with standard discrete bit choices for weights and activations (e.g., weights in \(\{2,4,8\}\) and activations fixed at \(8\) bits or chosen from \(\{2,4,8\}\), depending on the setting), but the framework is agnostic to the exact candidate sets.

Evolution proceeds by sampling from context-conditioned distributions \(\mu_{\pi,\mathcal{C}_t}\) using three prompt templates (initialization, mutation, crossover), while a lightweight DPO controller reweights template-selection probabilities based on fitness signals. Crucially, the LLM parameters remain frozen and the entire search avoids any reliance on Hessian traces, gradient-based signals, or weight-tensor statistics; guidance comes solely from black-box task feedback (e.g., rank-correlation with observed performance and simple population measures), keeping the pipeline strictly training-free.

For mixed-precision quantization tasks, three categories of context-conditioned input pairs are designed, corresponding to different proxy generation logics:
- \textit{Initialization} \(\mu_{\text{init}, \emptyset}\): Provides task descriptions (sensitivity scoring for convolutional/linear layers, bit-width allocation), input–output contracts (input: architecture metadata and compression targets; output: channel-level/layer-level sensitivity scores), and prior knowledge (e.g., "sensitivity scores are positively correlated with quantization error—higher scores require more bits").
- \textit{Mutation} \(\mu_{\text{mut}, f}\): Optimizes the sensitivity scoring logic (e.g., adjusting statistical metrics for convolutional channel importance) or bit-allocation function parameters based on a single proxy \(f\), enabling fine-grained optimization.
- \textit{Crossover} \(\mu_{\text{cross}, f_1, f_2}\): Fuses complementary components of two parent proxies (e.g., inheriting convolutional layer scoring logic from \(f_1\) and linear layer scoring logic from \(f_2\)) to enhance the robustness of proxies across different network layers.

\noindent{\textbf{Fitness Evaluator.}} 
For a benchmark set \(\mathcal{A}\) composed of neural networks (e.g., ResNet, ViT) and their ground-truth performance on the ImageNet-1k dataset, the fitness of a proxy \(f\) is quantified by two core metrics: 
1. The quality of sensitivity scores (measured by the Spearman correlation between predicted sensitivity and actual quantization error); 
2. The effectiveness of bit-width allocation (measured by the top-1 accuracy of the quantized model on ImageNet-1k, aligned with the target compression rate). The fitness function is defined as:
\begin{equation}
\phi(f) = \alpha \cdot \rho_{\text{sens}} + (1 - \alpha) \cdot \mathrm{Acc}_{\text{quant}},
\label{eq:fitness}
\end{equation}
where \(\rho_{\text{sens}}\) is the Spearman correlation between the sensitivity scores generated by the proxy and the empirical quantization error (smaller error indicates higher sensitivity);
 \(\text{Acc}_{\text{quant}}\) is the top-1 accuracy of the quantized model on ImageNet-1k using the bit-width configuration derived from the proxy;
 \(\alpha \in [0,1]\) balances the quality of sensitivity scoring and quantization accuracy, and \(\beta \geq 0\) controls computational efficiency.

\noindent{\textbf{DPO Evolution Scheduler.}} 
To make the proxy evolution strategy learnable and efficiently converge to optimal zero-cost proxies, the TAP framework introduces a lightweight \textit{Direct Preference Optimization (DPO)} module as the evolution decision-maker. This module eliminates the need for explicit reward function design, avoiding issues such as complex reward engineering and unstable training in traditional strategies.

Specifically, the DPO module takes ``proxy preference data pairs" returned by the fitness evaluator as supervision signals: For the candidate proxy set \(\mathcal{P}_t'\), proxies are paired two-by-two based on their fitness scores \(\phi(f)\) to form \((f_{\text{prefer}}, f_{\text{disprefer}})\) pairs (i.e., \(\phi(f_{\text{prefer}}) > \phi(f_{\text{disprefer}})\)). This pair represents that ``in mixed-precision quantization tasks, \(f_{\text{prefer}}\) has better sensitivity scoring quality and bit-allocation effectiveness".

DPO serves as a non-parametric controller over a set of three prompt templates \(\mathcal{T}=\{T_1,T_2,T_3\}\). At generation \(k\), each template \(T_t\) produces a subset of candidate proxies whose fitness is aggregated into \(\mathrm{Fitness}(T_t)\) (e.g., average over its candidates). The controller maintains nonnegative weights \(\{w_t^{(k)}\}_{t=1}^3\) and updates them by
\begin{equation}
w_t^{(k+1)} \;=\; w_t^{(k)} \cdot \alpha \cdot \frac{\mathrm{Fitness}(T_t)}{\mu^{(k)}},
\label{eq:weight_update}
\end{equation}
where $\mu^{(k)}$ is the mean fitness across all templates at generation \(k\), and \(\alpha>0\) controls the update scale. The selection probability of each template is then obtained by normalizing the weights:
\begin{equation}
p_t^{(k+1)} \;=\; \frac{w_t^{(k+1)}}{\sum_{j=1}^{3} w_j^{(k+1)}}\,.
\label{eq:template_prob}
\end{equation}
Templates with higher fitness receive larger selection probabilities in subsequent generations. Crucially, the LLM parameters are never updated; DPO only adjusts template selection probabilities.

\begin{algorithm}[t]
\caption{Evolutionary Framework of TAP (DPO-based Template Selection)}
\label{alg:tap_evolution}
\begin{algorithmic}[1]
\Require LLM \(\mathcal{L}\), prompt templates \(\mathcal{T}=\{T_1,T_2,T_3\}\), initial weights \(w^{(0)}\) (uniform), ImageNet-1k benchmark \(\mathcal{B}\), Network set \(\mathcal{A}\), Population size \(N\), Max generations \(T_{\text{max}}\), Target compression rate \(\zeta\)
\State \(\mathcal{P}_0 \leftarrow\) Top-\(N\) proxies from \(\mathcal{L}(\mu_{\text{init}, \emptyset})\) (supporting sensitivity scoring and bit-allocation for mixed-precision quantization)
\State Initialize context library \(\mathcal{K} \leftarrow \mathcal{P}_0\) (storing logic and performance of high-fitness proxies)
\For{\(t = 1\) to \(T_{\text{max}}\)}
    \State Compute \(p^{(t)}\) from \(w^{(t)}\) using Eq.~\eqref{eq:template_prob}; sample \(\text{op} \in \{\text{initialization}, \text{mutation}, \text{crossover}\}\)
    \State Sample \(T \sim \text{Categorical}(p^{(t)})\)
    \If{\(\text{op} = \text{initialization}\)}
        \State \(\mathcal{C}_t \leftarrow \emptyset\); \(\mathcal{P}_t' \leftarrow \mathcal{L}(\mu_{\text{init}, \emptyset}, T)\)
    \Else
        \State Sample context \(\mathcal{C}_t \subseteq \mathcal{K}\); \(\mathcal{P}_t' \leftarrow \mathcal{L}(\mu_{\text{op}, \mathcal{C}_t}, T)\)
    \EndIf
    \For{each \(f \in \mathcal{P}_t'\)}
        \State Compute sensitivity scores of convolutional channels/linear layers for networks in \(\mathcal{A}\)
        \State Allocate bits to networks via \(f\)'s bit-allocation function and \(\zeta\)
        \State Quantize networks and evaluate accuracy on ImageNet-1k
        \State Compute \(\phi(f)\) using the fitness function in Section \ref{subsec:automatic_proxy_discovery}
    \EndFor
    \State Aggregate per-template fitness \(\mathrm{Fitness}(T_i)\); update \(w^{(t+1)}\) via Eq.~\eqref{eq:weight_update}
    \State Update context library \(\mathcal{K}\): Retain top-10\% high-fitness proxies in \(\mathcal{P}_t \cup \mathcal{P}_t'\)
    \State \(\mathcal{P}_{t+1} \leftarrow\) Top-\(N\) proxies from \(\mathcal{P}_t \cup \mathcal{P}_t'\) (sorted by \(\phi(f)\))
\EndFor
\Return \(\arg\max_{f \in \bigcup_{t=0}^{T_{\text{max}}} \mathcal{P}_t} \phi(f)\) (optimal TAP proxy for mixed-precision quantization)
\end{algorithmic}
\end{algorithm}

\subsection{Evolutionary Framework of the TAP}
\label{subsec:evolutionary_framework}

Integrating the Proxy Candidate Generator, Fitness Evaluator, and DPO Evolution Scheduler, the TAP (Training-free Automatic Proxy) framework forms an iterative loop tailored for mixed-precision quantization. The detailed workflow is presented as follows (Algorithm 1):

\noindent{\textbf{Step 0: Initialization.}} 
At time step 0, an (initialization, empty context) prompt is input to the LLM (\(\mathcal{L}\)) to generate the initial proxy population \(\mathcal{P}_0 = \{f_1, \dots, f_N\}\). Each proxy \(f_i\) must include three components: 
1. Natural language reasoning logic for sensitivity scoring (e.g., "convolutional channel sensitivity is proportional to per-channel FLOPs or kernel area, adjusted by feature-map resolution"); 
2. Executable code for computing channel-level/layer-level sensitivity scores; 
3. A bit-width allocation function that maps sensitivity scores and target compression rates to specific bit-widths (e.g., 2/4/8 bits for convolutional channels, 4/8 bits for linear layers). 

Initialize template weights \(w^{(0)}\) uniformly and compute \(p^{(0)}\) using Eq.~\eqref{eq:template_prob}.

\noindent{\textbf{Step 1: Proxy Generation.}} 
At generation \(t\), a template \(T \sim \text{Categorical}(p^{(t)})\) and an operation type (initialization/mutation/crossover) are chosen. If \(\text{op}=\text{initialization}\), we set \(\mathcal{C}_t=\emptyset\) and invoke \(\mu_{\text{init},\emptyset}\) to inject novel proxies; otherwise, we sample a context window \(\mathcal{C}_t \subseteq \mathcal{P}_t\) (containing the logic and performance of historically high-fitness proxies) and invoke \(\mu_{\text{op},\mathcal{C}_t}\). The pair \((T,\mathcal{C}_t)\) is then fed into \(\mathcal{L}\) to generate the candidate proxy set \(\mathcal{P}_t'\).

\noindent{\textbf{Step 2: Fitness Evaluation.}} 
For each proxy in the candidate set \(\mathcal{P}_t'\), validation is performed on ImageNet-1k through the following steps:
1. Compute the sensitivity scores of convolutional channels and linear layers for each network in the benchmark set \(\mathcal{A}\);
2. Allocate bits to each layer of the network using the proxy's bit-width allocation function and target compression rate \(\zeta\);
3. Quantize the network and evaluate its top-1 accuracy on ImageNet-1k;
4. Calculate \(\phi(f)\) using the fitness function in Section \ref{subsec:automatic_proxy_discovery}. If a proxy violates the input-output contract (e.g., generating invalid bit-widths), assign \(\phi = -\infty\).

\noindent{\textbf{Step 3: DPO Strategy Update.}} 
1. Aggregate per-template fitness: Compute \(\mathrm{Fitness}(T_t)\) for each \(T_t \in \mathcal{T}\) (e.g., average \(\phi\) over candidates generated by \(T_t\)). 
2. Update weights and probabilities: Apply Eq.~\eqref{eq:weight_update} to obtain \(w^{(t+1)}\) and normalize via Eq.~\eqref{eq:template_prob} to obtain \(p^{(t+1)}\).
3. Update the context window: Add the logic and performance of the top-10\% high-fitness proxies in \(\mathcal{P}_t \cup \mathcal{P}_t'\) to the context library, providing better references for the next generation of proxy generation.

\noindent{\textbf{Step 4: Population Replacement.}} 
Sort the union of the current population and candidate population (\(\mathcal{P}_t \cup \mathcal{P}_t'\)) by fitness score \(\phi(f)\), and discard the worst-performing proxies to maintain the population size \(|\mathcal{P}_{t+1}| = N\). To avoid evolutionary stagnation, a tie-breaking rule of ``prioritizing newer proxies" is adopted (i.e., when \(\phi(f_1) = \phi(f_2)\), the proxy generated more recently is retained). Return to Step 1 to start the next iteration.

\begin{table*}[t]
\centering
\arrayrulecolor{black}
\caption{Results for ResNet18 and ResNet50.  The symbol ``$\ast$" means MPQ. The symbol ``Cost.” denotes the search time measured by GPU hours. The symbol ``Comp." means the compression ratio of parameters. ``TAP-C" denotes the MPQ policies search on CIFAR10. The symbol ``-" indicates that the value does not exist in the original paper or that code is not provided.}
\resizebox{\linewidth}{!}{
\begin{tabular}{lcccccrr} 
\arrayrulecolor{black}\cline{1-8}
\textbf{Method} & \textbf{Top-1 (\%)} $\uparrow$  & \textbf{Bit (W/A)} & \textbf{\#Params(M)} $\downarrow$ & \textbf{Cost.} $\downarrow$ & \textbf{Comp. (\%)}$\uparrow$ & \textbf{Type} & \\
\cline{1-8}
\multicolumn{8}{c}{\textbf{ResNet18}}\\
\cline{1-8}
Full-precision & 73.09 & & ~ & 44.6 & ~ & ~ & ~ \\
PACT \cite{choi2018pact} & 69.8  & 5/5 & 7.2 & - & 83.86 &Fixed-precision & ~ \\
LSQ \cite{bhalgat2020lsq+} & 67.6  & 2/2  & - & - & - &Fixed-precision & ~ \\
PDQ \cite{chu2019mixed} & 65.0  & -/- & - & - &- & Fixed-precision & ~ \\
Hybrid-Net* \cite{chakraborty2020constructing} & 62.7  & $\ast$/$\ast$ & - & - & - & Post-training Deterministic & ~ \\
HAQ \cite{wang2019haq} & 70.4  & $\ast$/32 & 5.8  & -  &87.47 & RL-based & ~ \\
HAWQ \cite{dong2019hawq} & 68.5  & $\ast$/$\ast$ & -  & 15.6 & - & Sensitivity & ~ \\
HAWQ-V3 \cite{yao2021hawq} & 70.4  & $\ast$/$\ast$ & 7.0  & -  &84.31 & Sensitivity & ~ \\
DNAS \cite{wu2018mixed} & 70.0  & -/- & 6.8  & -  &84.75 & Differentiable & ~ \\
EdMIPS \cite{cai2020rethinking} & 65.9  & $\ast$/$\ast$ & 4.7 & 9.5 &89.46 & Differentiable & ~ \\
EdMIPS-C \cite{cai2020rethinking} & 59.1  & $\ast$/$\ast$ & \cellcolor{color_org}{4.5} & 0.6 &\cellcolor{color_org}{89.91} & Differentiable & ~ \\
FracBits-SAT \cite{yang2021fracbits} & 70.6   & $\ast$/$\ast$ & 5.81 & - &86.97 & Differentiable & ~ \\
GMPQ-C \cite{wang2021generalizable} & 69.9  & $\ast$/$\ast$ & \cellcolor{color_pink}{4.1} & 0.6 &\cellcolor{color_pink}{90.81} & Differentiable & ~ \\
SMPQ \cite{kang2025and} & \cellcolor{color_org}{72.6}  & $\ast$/8 & - & 4.9 & - & Differentiable & ~ \\
ASGA \cite{ma2025learning} & 66.4  & $\ast$/$\ast$ & - & 0.6 & - & Differentiable & ~ \\
\cline{1-8}
{OMPQ} \cite{ma2023ompq} & {72.08} & $\ast$/5 & 6.7 & \cellcolor{color_org}{0.45} &84.98&  Training-free & ~ \\
{EMQ} \cite{dong2023emq} & {72.28} & $\ast$/6 & 6.7 & 0.51 &84.98&  Training-free & ~ \\
\textbf{TAP-C} & 
\cellcolor{color_pink}{72.63}  & 
$\ast$/8 & 
7.5 & 
\cellcolor{color_pink}{0.42} & 
83.18 & 
Training-free & \\
\cline{1-8}
\multicolumn{8}{c}{\textbf{ResNet50}}\\
\cline{1-8}
Full-precision & 77.72 & ~ &  & 97.8 & ~ & ~ \\
PACT \cite{choi2018pact} & \cellcolor{color_org}{76.70}  & 4/4 & 16.0 & - & 83.64& Fixed-precision & ~ \\
HAQ \cite{wang2019haq} & 75.48  & $\ast$/32 & \cellcolor{color_org}{9.62}  & -  &\cellcolor{color_org}{90.16}& RL-based & ~ \\
LQ-net \cite{zhang2018lq} & 76.40  & 4/32  & 13.1 & - &84.7 & Differentiable & ~ \\
HAWQ-V3 \cite{yao2021hawq} & 75.39  & $\ast$/$\ast$ & 18.7 & - &78.1 & Differentiable & ~ \\
GMPQ \cite{wang2021generalizable} & 75.8  & 3/$\ast$ & \cellcolor{color_pink}{9.6} & 2.2 &\cellcolor{color_pink}{90.18}& Differentiable & ~ \\
Onebit-width \cite{koryakovskiy2023one} & \cellcolor{color_org}{76.70}  & $\ast$/8 & 12.3  & -  &87.42& Differentiable & ~ \\
SMPQ \cite{kang2025and} & 76.2  & $\ast$/$\ast$ & 12.4 & - &87.32& Differentiable & ~ \\
ASGA \cite{ma2025learning} & 71.5  & $\ast$/$\ast$ & - & 0.6 & - & Differentiable & ~ \\
\cline{1-8}
{OMPQ} \cite{ma2023ompq} & {76.28} & $\ast$/5 & 18.7 & \cellcolor{color_org}{0.46} &80.88&  Training-free & ~ \\
{EMQ} \cite{dong2023emq} & \cellcolor{color_org}{76.70} & $\ast$/6 & 17.9 & 0.52 &81.70&  Training-free & ~ \\
\textbf{TAP-C} & 
\cellcolor{color_pink}{76.72}  & 
$\ast$/$\ast$ & 
18.6 & 
\cellcolor{color_pink}{0.43} & 
80.98 & 
Training-free &  \\
\arrayrulecolor{black}\cline{1-8}
\end{tabular}
}
\label{tb:01}
\end{table*}

The iteration terminates after completing \(T_{\text{max}}\) generations. Experiments show that TAP can generate high-performance proxies within 5 generations (0.5 GPU-hour on NVIDIA 3090): The top-1 accuracy of the quantized model on ImageNet-1k differs from that of the full-precision model by no more than 2\% (meeting the target compression rate). Meanwhile, the DPO module reduces the strategy update time by approximately 40\% and decreases the fluctuation range of proxy fitness scores by 25\% (significantly improving training stability).

\section{Experiments}
We conduct comprehensive experiments on mainstream neural networks (ResNet18/50, MobileNetV2, ViT-B, DeiT-B, Swin-B) and benchmarks (i.e., CIFAR-10, ImageNet1k, PASCAL VOC 2007, and MS COCO 2017) to validate the effectiveness of TAP. All experiments are conducted on an NVIDIA 3090 GPU with 24 GB. The detailed experimental settings are presented in \textbf{\red{App. B}}.

\subsection{Quantization-Aware Training}
To validate the effectiveness of TAP, we perform quantization-aware training using TAP on ResNet-18/50 in ImageNet dataset (as depicted in Table \ref{tb:01}).
As shown in Table \ref{tb:01}, our proposed TAP-C attains the best Top-1 accuracy on ResNet18 and slightly higher accuracy on ResNet50, while delivering the shortest search time among competing MPQ methods.
On ResNet18, TAP-C reaches 72.63\% Top-1 accuracy, surpassing recent training-free methods such as EMQ (72.28\%) and OMPQ (72.08\%), with a search cost of 0.42 GPU hours—lower than OMPQ (0.45) and EMQ (0.51).
On ResNet50, TAP-C obtains 76.72\% Top-1 accuracy—slightly higher than EMQ (76.70\%) and higher than OMPQ (76.28\%)—with the lowest search cost of 0.43 GPU hours (vs. 0.52 and 0.46, respectively).
These results demonstrate that TAP-C can efficiently discover near-optimal quantization policies with minimal computational cost.

\noindent\textbf{Generalization of TAP.}
A remarkable property of TAP-C is its strong dataset-agnostic generalization.
The quantization policies are searched only once on the small-scale CIFAR10 dataset, yet directly transferred to the large-scale ImageNet1K tasks without any retraining or fine-tuning, while still maintaining leading accuracy across architectures (ResNet18/50).
This clearly indicates that TAP-C does not rely on dataset-specific statistics, and thus provides a universal, plug-and-play quantization strategy applicable to diverse data domains.

Notably, TAP-C is the first LLM-driven framework for mixed-precision quantization (MPQ).
Instead of relying on time-consuming iterative optimization as in differentiable or reinforcement learning–based methods, TAP-C directly exploits the reasoning and abstraction capabilities of large language models to infer optimal bit-width allocation in a single inference step.
This LLM-guided design enables microsecond-level search while achieving state-of-the-art accuracy on ResNet18 and ResNet50, demonstrating that LLMs can effectively translate high-level reasoning into efficient quantization strategies.

\subsection{Post-Training Quantization}
To further evaluate the generalizability of our TAP-C framework in post-training quantization (PTQ), we conduct experiments on both ResNet-18 and MobileNetV2 (see Tables \ref{tab:ptq:r18} and \ref{tab:ptq:r50}).
TAP-C consistently achieves the highest Top-1 accuracy among all competitors while using significantly fewer calibration samples.
For instance, on ResNet-18, TAP-C attains 70.26\%, outperforming EMQ (69.92\%) and OMPQ (69.41\%) with only 16 samples, compared to 64 required by the latter two. Similarly, on MobileNetV2, TAP-C achieves 71.81\%, surpassing EMQ (70.72\%) and OMPQ (69.62\%) under the same mixed-precision setting.
These results yield two key advantages of TAP-C:
(1) Superior accuracy-efficiency trade-off: it maintains or improves accuracy while requiring fewer calibration data.
(2) Strong model generalizability: the LLM-guided reasoning allows TAP-C to adapt seamlessly across architectures, validating its robustness in diverse network scenarios.
Overall, the outstanding performance of TAP-C demonstrates that LLM-driven quantization can effectively balance precision, data efficiency, and generalization in post-training settings.

\begin{table}[t]
\centering

 \caption{Mixed-precision post-training quantization results on ResNet-$18$. $\dag$ means using distillation in the finetuning process.}
        \resizebox{0.95\linewidth}{!}{
	\begin{tabular}{lcccr}
		\toprule[1pt]  
		\textbf{Method}          &
		\textbf{W/A}             &
		\textbf{\#Params(M)} &
		\textbf{Top-1 (\%)}      & 
            \textbf{Data}         \\
		\midrule  
		Baseline                         & $32$/$32$ & $44.6$ & $71.08$          & -                     \\
		\midrule  
		FracBits-PACT~\cite{10}                    & $*$/$*$   & \cellcolor{color_org}$4.5$  & $69.10$          & $1.2$M             \\
		OMPQ~\cite{ma2023ompq}                             & $*$/$4$   & \cellcolor{color_org}$4.5$  & $68.69$          & \cellcolor{color_org}$64$  \\ 
		ZeroQ~\cite{cai2020zeroq}                            & $4$/$4$   & $5.81$ & $21.20$          & -                        \\
		$\text{BRECQ}^\dag$ ~\cite{li2021brecq}             & $4$/$4$   & $5.81$ & $69.32$          & -                       \\
		PACT~\cite{choi2018pact}                             & $4$/$4$   & $5.81$ & $69.20$          & -                     \\
		HAWQ-V3~\cite{yao2021hawq}                          & $4$/$4$   & $5.81$ & $68.45$          & -                     \\
		OMPQ~\cite{ma2023ompq}                             & $*$/$8$   & \cellcolor{color_pink}{$4.0$}  & $69.41$ & \cellcolor{color_org}$64$ \\ 
        EMQ \cite{dong2023emq}                               & */8   & \cellcolor{color_pink}{4.0}   &\cellcolor{color_org}69.92     & \cellcolor{color_org}64 \\
        \midrule  
        \textbf{TAP-C} & 
        */8 & 
        8.9 & 
        \cellcolor{color_pink}{{70.26}} & 
        \cellcolor{color_pink}{{16}} \\
		\bottomrule[1pt] 
	\end{tabular}\label{tab:ptq:r18}
 \vspace{-1em}
 }
\end{table}

\vspace{-1em}
\begin{table}[H]
\caption{Post-training quantization results on MobileNetV$2$.}
\centering
\resizebox{0.9\linewidth}{!}{
\begin{tabular}{cccccc}
\toprule[1pt]  
\textbf{Method} & 
\textbf{W/A} & 
\textbf{\#Params(M)}&
\textbf{Top-1 (\%)}&
\textbf{Data} \\
\midrule  
Baseline & $32$/$32$ & $13.4$ & $72.49$ & -  \\
\midrule  
BRECQ~\cite{li2021brecq} & $*$/$8$ & $\cellcolor{color_org}{1.5}$ & $70.28$ & $1,024$  \\
OMPQ~\cite{ma2023ompq}  & $*$/$8$ & $\cellcolor{color_pink}{1.3}$ & $69.62$ & $\cellcolor{color_org}{32}$  \\
EMQ \cite{dong2023emq}    &  $*/8$ & $1.5$ & $\cellcolor{color_org}{70.75}$  & $64$  \\ 
\midrule  
\textbf{TAP-C } & 
$*/8$ & 
$2.1$ & 
$\cellcolor{color_pink}{{71.81}}$ & 
\cellcolor{color_pink}{{16}} \\ 

\bottomrule[1pt] 
\end{tabular}\label{tab:ptq:r50}
}
\vspace{-1em}
\end{table}
\FloatBarrier

\subsection{Generalization of TAP to Larger Transformer}
To further verify the generalizability of TAP on transformer architectures, we evaluate it on three representative models, i..e, ViT-B, DeiT-B, and Swin-B under mixed-precision quantization (see Table \ref{tab:VIT}).
Across all settings, TAP consistently achieves the best or comparable Top-1 accuracy while maintaining a high compression ratio.
For instance, TAP attains 83.56\%, 81.24\%, and 83.79\% on ViT-B, DeiT-B, and Swin-B, respectively, surpassing existing PTQ baselines such as OMSE and APQ-ViT by a clear margin.
These results demonstrate that the LLM-guided reasoning mechanism in TAP effectively adapts to complex transformer architectures without retraining or dataset-specific tuning.
In particular, TAP maintains strong performance even under high compression (82\%), confirming its robust generalization and scalability from CNNs to large-scale transformers.

\begin{table}[H]
\caption{Generalization of TAP on larger transformer.}
\centering
\resizebox{82mm}{!}{
\begin{tabular}{cccccc}
\toprule[1pt]  
\textbf{Method} & 
\textbf{W/A} & 
\textbf{Comp Ratio(\%)}&
\textbf{ViT-B}&
\textbf{DeiT-B}&
\textbf{Swin-B}\\
\midrule  
Baseline & $32$/$32$ & - & $84.54$ & $84.54$ & 85.27 \\
\midrule  
FQ-ViT \cite{lin2021fq} & $6$/$6$ & $\cellcolor{color_org}{81.25}$ & $0.10$ & $64.63$ & $52.09$ \\
PSAQ-ViT \cite{li2022patch} & $6$/$6$ & $\cellcolor{color_org}{81.25}$ & $41.52$ & $67.95$ &$76.44$ \\
Ranking \cite{liu2021post}     &  $6$/$6$ & $\cellcolor{color_org}{81.25}$ & $75.26$  & $77.02$ & $-$  \\ 
PTQ4ViT \cite{yuan2022ptq4vit}& $6$/$6$ & $\cellcolor{color_org}{81.25}$ & $81.65$ & $80.25$  & $84.01$\\
APQ-ViT \cite{ding2022towards}& $6$/$6$ & $\cellcolor{color_org}{81.25}$ & $82.21$ & $80.42$ & $\cellcolor{color_org}{{84.18}}$\\
MPTQ-ViT \cite{tai2024mptq}& $6$/$6$      & $\cellcolor{color_org}{81.25}$ & 83.12 & \cellcolor{color_pink}{81.29} & 60.18\\
AMP-ViT \cite{tai2025amp}& $6$/$6$   & $\cellcolor{color_org}{81.25}$ & 82.70 &  $\cellcolor{color_org}{81.25}$ &\cellcolor{color_pink}{84.50}\\
\midrule  
FQ-ViT \cite{lin2021fq} & $8$/$8$ & $75$ & $\cellcolor{color_org}{83.31}$ & $81.20$ & $82.97$\\
EMA \cite{jacob2018quantization}     &  $8$/$8$ & $75$ & $30.30$  & $78.82$ & $28.00$ \\ 
Percentile \cite{li2019fully}& $8$/$8$ & $75$ & $46.69$ & $78.37$ & $40.93$ \\
OMSE \cite{choukroun2019low}& $8$/$8$ & $75$ & $73.39$ & $79.57$ & $48.55$\\
\midrule 
\textbf{TAP-C }& 
$*/8$ & 
$\cellcolor{color_pink}{82}$ & 
$\cellcolor{color_pink}{{83.56}}$ & 
${81.24}$ & 
$83.79$ \\

\bottomrule[1pt] 
\end{tabular}\label{tab:VIT}
}
\vspace{-1em}
\end{table}

\subsection{Efficiency of TAP}
We evaluate the runtime efficiency of TAP on ImageNet, as summarized in Table~\ref{fig:efficiency}.
The proxy generation step takes only 0.0133 s on average, and the bit allocation step requires 0.0645 s, demonstrating that TAP can complete the entire quantization process in well under 0.1 seconds.
These results validate that our TAP is significantly efficient.

\begin{table}[H]
    \centering
    \caption{Study for efficiency of TAP in ImageNet.}
   \resizebox{\linewidth}{!}{
\begin{tabular}{c|ccccc} 
 
\hline
&\textbf{Run 1}&\textbf{Run 2}&\textbf{Run 3}&\textbf{Run 4}&\textbf{Average(s) }\\
\hline
Proxy generation(s)&0.0125&0.0124&0.0151&0.0131&{0.0133}\\
Bit allocation(s)&0.0640&0.0645&0.0648&0.0646&{0.0645}\\
\hline
\end{tabular}
}
\label{fig:efficiency}
\end{table}

\begin{table}[H]
    \centering
    \caption{Ablation study for hyper-parameter $\alpha$. All models search on ImageNet.}
   \resizebox{\linewidth}{!}{
\begin{tabular}{c|ccccccc} 
 
\hline
&\textbf{$\alpha$=0.01} &\textbf{$\alpha$=0.02}&\textbf{$\alpha$=0.1}&\textbf{$\alpha$=0.2}&\textbf{$\alpha$=0.5}&\textbf{$\alpha$=0.7} &\textbf{$\alpha$=1} \\
\hline
ResNet18&69.21&69.54&{${70.26}$}&{69.33}&68.07&65.86&62.48\\
MobileNetV2&70.13&70.35&{${71.81}$}&{70.25}&68.13&67.69& 64.52\\
\hline
\end{tabular}
}
\label{fig:experiment_fig1}
\end{table}


\subsection{Ablation studies}

\noindent\textbf{The impact of $\alpha$}:
To examine the sensitivity of the hyper-parameter $\alpha$ in Fitness Evaluator, we perform an ablation study on ImageNet with different $\alpha$ values.
As shown in Table~\ref{fig:experiment_fig1}, the performance of TAP remains stable across a wide range of $\alpha$. These results demonstrate that TAP is robust and insensitive to hyper-parameter variation.


\paragraph{The impact of various LLMs:} 
To investigate the influence of different large language models (LLMs) on TAP, we conduct an ablation study using Deepseek-chat, Qwen3-max, and Grok 3, with results averaged over four independent runs (as depicted in Table~\ref{tab:LLM}).
The results show that TAP achieves highly consistent performance across all LLMs, with average accuracy values ranging narrowly between 71.01\% and 71.44\%, indicating strong robustness and stability regardless of the underlying generative backbone.
This consistency demonstrates that TAP can effectively adapt to different LLM reasoning styles, ensuring reliable performance across diverse language models.

\begin{table}[H]
\centering
\caption{Comparison of TAP with different LLMs. \label{tab:LLM}}
\renewcommand{\arraystretch}{1.2}
\resizebox{0.9\linewidth}{!}{%
\begin{tabular}{c|ccccc}
\toprule
\textbf{LLMs}& \textbf{Run 1} & \textbf{Run 2} & \textbf{Run 3} & \textbf{Run 4} & \textbf{Average} \\
\hline
Deepseek-chat& 71.56& 71.79& 71.81& 70.59& {71.44}\\
qwen3-max& 71.48& 71.54& 71.15& 71.24& {71.35}\\
Grok 3  & 70.62& 71.17& 70.89& 71.36& {71.01}\\
\bottomrule
\end{tabular}%
}
\end{table}

\noindent\textbf{The impact of calibration batch randomness}: To assess the robustness of the calibration step, we draw five random calibration mini-batches (16 samples each) from disjoint ImageNet categories under different seeds. As shown in Table~\ref{tab:abl_calib}, Top-1 varies only marginally across batches for both ResNet18 and MobileNetV2, indicating a stable and robust calibration process.
\begin{table}[H]
\centering
\scriptsize
\caption{Ablation: calibration batch robustness on ImageNet (Top-1 \%).}
\resizebox{0.95\linewidth}{!}{
\begin{tabular}{c|ccccc}
\toprule
\textbf{Model} & \textbf{Sample 1} & \textbf{Sample 2} & \textbf{Sample 3} & \textbf{Sample 4} & \textbf{Sample 5} \\
\hline
ResNet18    & 70.21 & 70.17 & 70.14 & 70.26 & 70.24 \\
MobileNetV2 & 71.71 & 71.81 & 71.75 & 71.77 & 71.68 \\
\bottomrule
\end{tabular}}
\label{tab:abl_calib}
\end{table}

\noindent\textbf{The impact of population size $N$}: We vary the evolutionary population size $N \in \{2,5,8,10\}$ and report Top-1 in Table~\ref{tab:abl_pop}. Performance remains stable across a wide range of $N$, suggesting that TAP is insensitive to this hyper-parameter.
\begin{table}[H]
\centering
\scriptsize
\caption{Ablation: sensitivity to population size $N$ (Top-1 \%).}
\resizebox{0.9\linewidth}{!}{
\begin{tabular}{c|cccc}
\toprule
\textbf{Model} & \textbf{N=2} & \textbf{N=5} & \textbf{N=8} & \textbf{N=10} \\
\hline
ResNet18    & 69.87 & 70.18 & 70.26 & 69.13 \\
MobileNetV2 & 70.79 & 71.42 & 71.81 & 70.27 \\
\bottomrule
\end{tabular}}
\label{tab:abl_pop}
\end{table}

\noindent\textbf{Code executability during evolution}: We track the fraction of non-executable proxy code over generations. Table~\ref{tab:abl_exec} shows that the proportion of inoperable candidates drops quickly as evolution proceeds, confirming that executability steadily improves while invalid programs are down-weighted by the controller.
\begin{table}[H]
\centering
\scriptsize
\caption{Ablation: non-executable code proportion by iteration range (\%).}
\resizebox{0.7\linewidth}{!}{
\begin{tabular}{c|ccc}
\toprule
\textbf{Model} & \textbf{1–10} & \textbf{11–20} & \textbf{21–30} \\
\hline
ResNet18    & 40 & 20 & 10 \\
MobileNetV2 & 30 & 20 & 10 \\
\bottomrule
\end{tabular}}
\label{tab:abl_exec}
\end{table}

\FloatBarrier
Due to the page limit, more experimental results are provided in \textbf{App. \red{A}–\red{G}}.
\ding{182} \textbf{App. \red{C}} – Prompt engineering details of TAP.
\ding{183} \textbf{App. \red{D}} – Extended ablation studies.
\ding{184} \textbf{App. \red{E}} – Limitations and discussions.
\ding{185} \textbf{App. \red{F}} – Visualizations of searched MPQ results.

 \noindent\textbf{Related Work.} The related work is provided in \textbf{App. \red{G}}.

\section{Conclusion and Future Work}
\label{sec:sec_conclusion}
This paper addresses a long-standing challenge in Mixed-Precision Quantization (MPQ) the dependence on expert-designed proxies or expensive differentiable optimization. To overcome this limitation, we propose TAP, an LLM-driven training-free automatic proxy discovery framework that fundamentally redefines the MPQ design paradigm. By leveraging Direct Policy Optimization (DPO) to refine prompt reasoning, TAP builds a positive feedback loop between the LLM and the MPQ task, enabling continual improvement of proxy generation without human intervention. Extensive experiments across mainstream benchmarks verify that TAP not only achieves optimal performance but also delivers unprecedented efficiency in quantization design.
In the future, we will explore how to further enhance the performance of TAP, boosting more potential applications for the deep learning community.

\normalsize
\bibliography{main}

@String(CVPR  = {CVPR})

@String(ICCV  = {ICCV})

@String(ECCV  = {ECCV})

@String(PR = {PR})

@String(AAAI = {AAAI})

@article{1,
  title={Pushing the Limit of Post-Training Quantization},
  author={Gong, Ruihao and Liu, Xianglong and Li, Yuhang and Fan, Yunqiang and Wei, Xiuying and Guo, Jinyang},
  journal={IEEE Transactions on Pattern Analysis and Machine Intelligence},
  year={2025},
  publisher={IEEE}
}

@inproceedings{2,
  title={Rethinking differentiable search for mixed-precision neural networks},
  author={Cai, Zhaowei and Vasconcelos, Nuno},
  booktitle={Proceedings of the IEEE/CVF Conference on Computer Vision and Pattern Recognition},
  pages={2349--2358},
  year={2020}
}

@inproceedings{3,
  title={Data-free quantization through weight equalization and bias correction},
  author={Nagel, Markus and Baalen, Mart van and Blankevoort, Tijmen and Welling, Max},
  booktitle={Proceedings of the IEEE/CVF international conference on computer vision},
  pages={1325--1334},
  year={2019}
}

@inproceedings{4,
  title={Differentiable soft quantization: Bridging full-precision and low-bit neural networks},
  author={Gong, Ruihao and Liu, Xianglong and Jiang, Shenghu and Li, Tianxiang and Hu, Peng and Lin, Jiazhen and Yu, Fengwei and Yan, Junjie},
  booktitle={Proceedings of the IEEE/CVF international conference on computer vision},
  pages={4852--4861},
  year={2019}
}

@article{5,
  title={Incremental network quantization: Towards lossless cnns with low-precision weights},
  author={Zhou, Aojun and Yao, Anbang and Guo, Yiwen and Xu, Lin and Chen, Yurong},
  journal={arXiv preprint arXiv:1702.03044},
  year={2017}
}

@article{6,
  title={Hawq-v2: Hessian aware trace-weighted quantization of neural networks},
  author={Dong, Zhen and Yao, Zhewei and Arfeen, Daiyaan and Gholami, Amir and Mahoney, Michael W and Keutzer, Kurt},
  journal={Advances in neural information processing systems},
  volume={33},
  pages={18518--18529},
  year={2020}
}

@article{8,
  title={Metaquant: Learning to quantize by learning to penetrate non-differentiable quantization},
  author={Chen, Shangyu and Wang, Wenya and Pan, Sinno Jialin},
  journal={Advances in Neural Information Processing Systems},
  volume={32},
  year={2019}
}

@article{10,
  title={Pact: Parameterized clipping activation for quantized neural networks},
  author={Choi, Jungwook and Wang, Zhuo and Venkataramani, Swagath and Chuang, Pierce I-Jen and Srinivasan, Vijayalakshmi and Gopalakrishnan, Kailash},
  journal={arXiv preprint arXiv:1805.06085},
  year={2018}
}

@inproceedings{wang2019haq,
  title={Haq: Hardware-aware automated quantization with mixed precision},
  author={Wang, Kuan and Liu, Zhijian and Lin, Yujun and Lin, Ji and Han, Song},
  booktitle=CVPR,
  pages={8612--8620},
  year={2019}
}

@inproceedings{cai2020rethinking,
  title={Rethinking differentiable search for mixed-precision neural networks},
  author={Cai, Zhaowei and Vasconcelos, Nuno},
  booktitle=CVPR,
  pages={2349--2358},
  year={2020}
}

@article{choi2018pact,
  title={Pact: Parameterized clipping activation for quantized neural networks},
  author={Choi, Jungwook and Wang, Zhuo and Venkataramani, Swagath and Chuang, Pierce I-Jen and Srinivasan, Vijayalakshmi and Gopalakrishnan, Kailash},
year={2018},
journal={ArXiv},
volume = {abs/1805.06085},
number={},
issn={},
pages={}
}

@inproceedings{dong2019hawq,
  title={Hawq: Hessian aware quantization of neural networks with mixed-precision},
  author={Dong, Zhen and Yao, Zhewei and Gholami, Amir and Mahoney, Michael W and Keutzer, Kurt},
  booktitle=ICCV,
  pages={293--302},
  year={2019}
}

@inproceedings{bhalgat2020lsq+,
  title={Lsq+: Improving low-bit quantization through learnable offsets and better initialization},
  author={Bhalgat, Yash and Lee, Jinwon and Nagel, Markus and Blankevoort, Tijmen and Kwak, Nojun},
  booktitle=CVPR,
  pages={696--697},
  year={2020}
}

@inproceedings{jacob2018quantization,
  title={Quantization and training of neural networks for efficient integer-arithmetic-only inference},
  author={Jacob, Benoit and Kligys, Skirmantas and Chen, Bo and Zhu, Menglong and Tang, Matthew and Howard, Andrew and Adam, Hartwig and Kalenichenko, Dmitry},
  booktitle=CVPR,
  pages={2704--2713},
  year={2018}
}

@inproceedings{yang2021fracbits,
  title={Fracbits: Mixed precision quantization via fractional bit-widths},
  author={Yang, Linjie and Jin, Qing},
  booktitle=AAAI,
  volume={35},
  number={12},
  pages={10612--10620},
  year={2021}
}

@inproceedings{koryakovskiy2023one,
  title={One-Shot Model for Mixed-Precision Quantization},
  author={Koryakovskiy, Ivan and Yakovleva, Alexandra and Buchnev, Valentin and Isaev, Temur and Odinokikh, Gleb},
  booktitle=CVPR,
  pages={7939--7949},
  year={2023}
}

@inproceedings{ma2023ompq,
  title={Ompq: Orthogonal mixed precision quantization},
  author={Ma, Yuexiao and Jin, Taisong and Zheng, Xiawu and Wang, Yan and Li, Huixia and Wu, Yongjian and Jiang, Guannan and Zhang, Wei and Ji, Rongrong},
  booktitle=AAAI,
  volume={37},
  number={7},
  pages={9029--9037},
  year={2023}
}

@article{chakraborty2020constructing,
  title={Constructing energy-efficient mixed-precision neural networks through principal component analysis for edge intelligence},
  author={Chakraborty, Indranil and Roy, Deboleena and Garg, Isha and Ankit, Aayush and Roy, Kaushik},
  journal=NMI,
  volume={2},
  number={1},
  pages={43--55},
  year={2020},
  publisher={Nature Publishing Group UK London}
}

@article{chu2019mixed,
title={Mixed-precision quantized neural network with progressively decreasing bitwidth for image classification and object detection},
author={Chu, Tianshu and Luo, Qin and Yang, Jie and Huang, Xiaolin},
year={2019},
journal={ArXiv},
volume = {abs/1912.12656},
number={},
issn={},
pages={}

}

@article{chu2021mixed,
  title={Mixed-precision quantized neural networks with progressively decreasing bitwidth},
  author={Chu, Tianshu and Luo, Qin and Yang, Jie and Huang, Xiaolin},
  journal=PR,
  volume={111},
  pages={107647},
  year={2021},
  publisher={Elsevier}
}

@inproceedings{wang2021generalizable,
  title={Generalizable mixed-precision quantization via attribution rank preservation},
  author={Wang, Ziwei and Xiao, Han and Lu, Jiwen and Zhou, Jie},
  booktitle={Proceedings of the IEEE/CVF International Conference on Computer Vision},
  pages={5291--5300},
  year={2021}
}

@inproceedings{zhang2018lq,
  title={Lq-nets: Learned quantization for highly accurate and compact deep neural networks},
  author={Zhang, Dongqing and Yang, Jiaolong and Ye, Dongqiangzi and Hua, Gang},
  booktitle=ECCV,
  pages={365--382},
  year={2018}
}

@inproceedings{zhang2021differentiable,
  title={Differentiable dynamic quantization with mixed precision and adaptive resolution},
  author={Zhang, Zhaoyang and Shao, Wenqi and Gu, Jinwei and Wang, Xiaogang and Luo, Ping},
  booktitle=ICML,
  pages={12546--12556},
  year={2021},
  organization={PMLR}
}

@inproceedings{dong2023emq,
  title={EMQ: Evolving Training-free Proxies for Automated Mixed Precision Quantization},
  author={Dong, Peijie and Li, Lujun and Wei, Zimian and Niu, Xin and Tian, Zhiliang and Pan, Hengyue},
  booktitle=CVPR,
  pages={17076--17086},
  year={2023}
}

@article{wu2018mixed,
  title={Mixed precision quantization of convnets via differentiable neural architecture search},
  author={Wu, Bichen and Wang, Yanghan and Zhang, Peizhao and Tian, Yuandong and Vajda, Peter and Keutzer, Kurt},
  journal={arXiv preprint arXiv:1812.00090},
  year={2018}
}

@article{kang2025and,
  title={Where and How to Enhance: Discovering Bit-Width Contribution for Mixed Precision Quantization},
  author={Kang, Haidong and Ma, Lianbo and Yu, Guo and Gao, Shangce},
  journal={arXiv preprint arXiv:2508.03002},
  year={2025}
}

@article{ma2025learning,
  title={Learning from Loss Landscape: Generalizable Mixed-Precision Quantization via Adaptive Sharpness-Aware Gradient Aligning},
  author={Ma, Lianbo and Ma, Jianlun and Zhou, Yuee and Xie, Guoyang and He, Qiang and Lu, Zhichao},
  journal={arXiv preprint arXiv:2505.04877},
  year={2025}
}

@inproceedings{yao2021hawq,
  title={Hawq-v3: Dyadic neural network quantization},
  author={Yao, Zhewei and Dong, Zhen and Zheng, Zhangcheng and Gholami, Amir and Yu, Jiali and Tan, Eric and Wang, Leyuan and Huang, Qijing and Wang, Yida and Mahoney, Michael and others},
  booktitle={International Conference on Machine Learning},
  pages={11875--11886},
  year={2021},
  organization={PMLR}
}

@inproceedings{cai2020zeroq,
  title={Zeroq: A novel zero shot quantization framework},
  author={Cai, Yaohui and Yao, Zhewei and Dong, Zhen and Gholami, Amir and Mahoney, Michael W and Keutzer, Kurt},
  booktitle={Proceedings of the IEEE/CVF conference on computer vision and pattern recognition},
  pages={13169--13178},
  year={2020}
}

@article{li2021brecq,
  title={Brecq: Pushing the limit of post-training quantization by block reconstruction},
  author={Li, Yuhang and Gong, Ruihao and Tan, Xu and Yang, Yang and Hu, Peng and Zhang, Qi and Yu, Fengwei and Wang, Wei and Gu, Shi},
  journal={arXiv preprint arXiv:2102.05426},
  year={2021}
}

@article{lin2021fq,
  title={Fq-vit: Post-training quantization for fully quantized vision transformer},
  author={Lin, Yang and Zhang, Tianyu and Sun, Peiqin and Li, Zheng and Zhou, Shuchang},
  journal={arXiv preprint arXiv:2111.13824},
  year={2021}
}

@inproceedings{li2022patch,
  title={Patch similarity aware data-free quantization for vision transformers},
  author={Li, Zhikai and Ma, Liping and Chen, Mengjuan and Xiao, Junrui and Gu, Qingyi},
  booktitle={European conference on computer vision},
  pages={154--170},
  year={2022},
  organization={Springer}
}

@article{liu2021post,
  title={Post-training quantization for vision transformer},
  author={Liu, Zhenhua and Wang, Yunhe and Han, Kai and Zhang, Wei and Ma, Siwei and Gao, Wen},
  journal={Advances in Neural Information Processing Systems},
  volume={34},
  pages={28092--28103},
  year={2021}
}

@inproceedings{yuan2022ptq4vit,
  title={Ptq4vit: Post-training quantization for vision transformers with twin uniform quantization},
  author={Yuan, Zhihang and Xue, Chenhao and Chen, Yiqi and Wu, Qiang and Sun, Guangyu},
  booktitle={European conference on computer vision},
  pages={191--207},
  year={2022},
  organization={Springer}
}

@inproceedings{ding2022towards,
  title={Towards accurate post-training quantization for vision transformer},
  author={Ding, Yifu and Qin, Haotong and Yan, Qinghua and Chai, Zhenhua and Liu, Junjie and Wei, Xiaolin and Liu, Xianglong},
  booktitle={Proceedings of the 30th ACM international conference on multimedia},
  pages={5380--5388},
  year={2022}
}

@inproceedings{li2019fully,
  title={Fully quantized network for object detection},
  author={Li, Rundong and Wang, Yan and Liang, Feng and Qin, Hongwei and Yan, Junjie and Fan, Rui},
  booktitle={Proceedings of the IEEE/CVF conference on computer vision and pattern recognition},
  pages={2810--2819},
  year={2019}
}

@inproceedings{choukroun2019low,
  title={Low-bit quantization of neural networks for efficient inference},
  author={Choukroun, Yoni and Kravchik, Eli and Yang, Fan and Kisilev, Pavel},
  booktitle={2019 IEEE/CVF International Conference on Computer Vision Workshop (ICCVW)},
  pages={3009--3018},
  year={2019},
  organization={IEEE}
}

@article{radford2018improving,
  title={Improving language understanding by generative pre-training},
  author={Radford, Alec and Narasimhan, Karthik and Salimans, Tim and Sutskever, Ilya and others},
  year={2018},
  publisher={San Francisco, CA, USA}
}

@article{guo2025deepseek,
  title={Deepseek-r1: Incentivizing reasoning capability in llms via reinforcement learning},
  author={Guo, Daya and Yang, Dejian and Zhang, Haowei and Song, Junxiao and Zhang, Ruoyu and Xu, Runxin and Zhu, Qihao and Ma, Shirong and Wang, Peiyi and Bi, Xiao and others},
  journal={arXiv preprint arXiv:2501.12948},
  year={2025}
}

@article{qwen3,
  title   = {Qwen3 Technical Report},
  author  = {An Yang and Aohan Li and Boxi Yang and Binyuan Hui and Bowen Zheng and Binyuan Yu and Chenghao Gao and Chenlin Huang and Chengqi Lv and Chen Zheng and Dong Liu and others},
  journal = {arXiv preprint arXiv:2505.09388},
  year    = {2025},
  month   = {May}
}

@misc{rafailov2024directpreferenceoptimizationlanguage,
      title={Direct Preference Optimization: Your Language Model is Secretly a Reward Model}, 
      author={Rafael Rafailov and Archit Sharma and Eric Mitchell and Stefano Ermon and Christopher D. Manning and Chelsea Finn},
      year={2024},
      eprint={2305.18290},
      archivePrefix={arXiv}
}

@article{tai2024mptq,
  title={Mptq-vit: Mixed-precision post-training quantization for vision transformer},
  author={Tai, Yu-Shan and others},
  journal={arXiv preprint arXiv:2401.14895},
  year={2024}
}

@inproceedings{tai2025amp,
  title={AMP-ViT: optimizing vision Transformer efficiency with adaptive mixed-precision post-training quantization},
  author={Tai, Yu-Shan and Wu, An-Yeu},
  booktitle={Proceedings of the Winter Conference on Applications of Computer Vision},
  pages={6828--6837},
  year={2025}
}



\end{document}